\def\year{2022}\relax
\documentclass[letterpaper]{article} % DO NOT CHANGE THIS
\usepackage{aaai22}  % DO NOT CHANGE THIS
\usepackage{times}  % DO NOT CHANGE THIS
\usepackage{helvet}  % DO NOT CHANGE THIS
\usepackage{courier}  % DO NOT CHANGE THIS
\usepackage[hyphens]{url}  % DO NOT CHANGE THIS
\usepackage{hyperref}
\usepackage{graphicx} % DO NOT CHANGE THIS
\usepackage{natbib}  % DO NOT CHANGE THIS AND DO NOT ADD ANY OPTIONS TO IT
\usepackage{caption} % DO NOT CHANGE THIS AND DO NOT ADD ANY OPTIONS TO IT
\usepackage{multirow}
\usepackage{subcaption}
\DeclareCaptionStyle{ruled}{labelfont=normalfont,labelsep=colon,strut=off} % DO NOT CHANGE THIS
\usepackage{algorithm}
\usepackage{algorithmic}
\usepackage{amsmath,amssymb}
\usepackage{newfloat}
\usepackage{listings}
\floatstyle{ruled}
\newfloat{listing}{tb}{lst}{}
\floatname{listing}{Listing}
\title{Graph augmented Deep Reinforcement Learning in the GameRLand3D environment}
\author{
    Edward Beeching,\textsuperscript{\rm 1, \rm 2}
    Maxim Peter \textsuperscript{\rm 1}
    Philippe Marcotte \textsuperscript{\rm1}
    Jilles Debangoye, \textsuperscript{\rm 2}
    Olivier Simonin \textsuperscript{\rm 2}
    Joshua Romoff \textsuperscript{\rm 1}
    Christian Wolf\textsuperscript{\rm 3}
}
\affiliations{
    \textsuperscript{\rm 1} Ubisoft La Forge, Montreal\\
    \textsuperscript{\rm 2} INRIA Chroma team, CITI Laboratory. INSA-Lyon, France. \\
    \textsuperscript{\rm 3} Université de Lyon, INSA-Lyon, LIRIS, CNRS, France.\\
}

\begin{document}
\maketitle

\begin{abstract}
\begin{quote}
We address planning and navigation in challenging 3D video games featuring  maps with disconnected regions reachable by agents using special actions. In this setting, classical symbolic planners are not applicable or difficult to adapt. We introduce a hybrid technique combining a low level policy trained with reinforcement learning and a graph based high level classical planner. In addition to providing human-interpretable paths, the approach improves the generalization performance of an end-to-end approach in unseen maps, where it achieves a 20\% absolute increase in success rate over a recurrent end-to-end agent on a point to point navigation task in yet unseen large-scale maps of size 1km$\times$1km. In an in-depth experimental study, we quantify the limitations of end-to-end Deep RL approaches in vast environments and we also introduce ``GameRLand3D'', a new benchmark and soon to be released environment can generate complex procedural 3D maps for navigation tasks.
%An overview video is available here: \href{https://youtu.be/H0WAHvEeVyc}.
\end{quote}
\end{abstract}
\section{Introduction}
\noindent Long term planning and navigation is an important component when looking to realistically control a non-player character (NPC) in a video game. Whether the NPC is player facing or used for automated testing to find bugs and exploits, the ability to navigate in a complex environment is paramount and is the foundation of every NPC.
In the field of video games the Navigation Mesh (NavMesh) \cite{snook2000navmesh} has become the ubiquitous approach for planning and navigation. In recent years, the sheer scale of video game environments have highlighted the limitations of the NavMesh, particularly in games with complex navigation abilities such as double jumping, teleportation, jump pads, grappling hooks, and wall runs. 
Given the successes of Deep Reinforcement Learning (RL) in domains of navigation \cite{mirowski2016learning, jaderberg2017unreal, jaderberg2019human}, board games \cite{tesauro1994td, silver2016mastering, silver2017mastering} and Atari video games \cite{mnih2013playing, hessel2018rainbow, hafner2020mastering}, recent works \cite{ijcai2021-294} have looked to exploit these approaches for navigation in large 3D video game environments. These works have shown promise in smaller game worlds, but have highlighted that as environments scale to hundreds of thousands of square metres, the limitations of end-to-end learning-based methods are encountered. 

%In addition to the release of the open source GameRLand3D environment, we also introduce a new hybrid navigation method, which combines a recurrent baseline agent with a graph-based classical planner. The resulting hierarchical planner combines low level way-point-based control and navigation with interpretable high level plans.
\begin{figure}[t]
    \centering
    \includegraphics[width=0.5\textwidth]{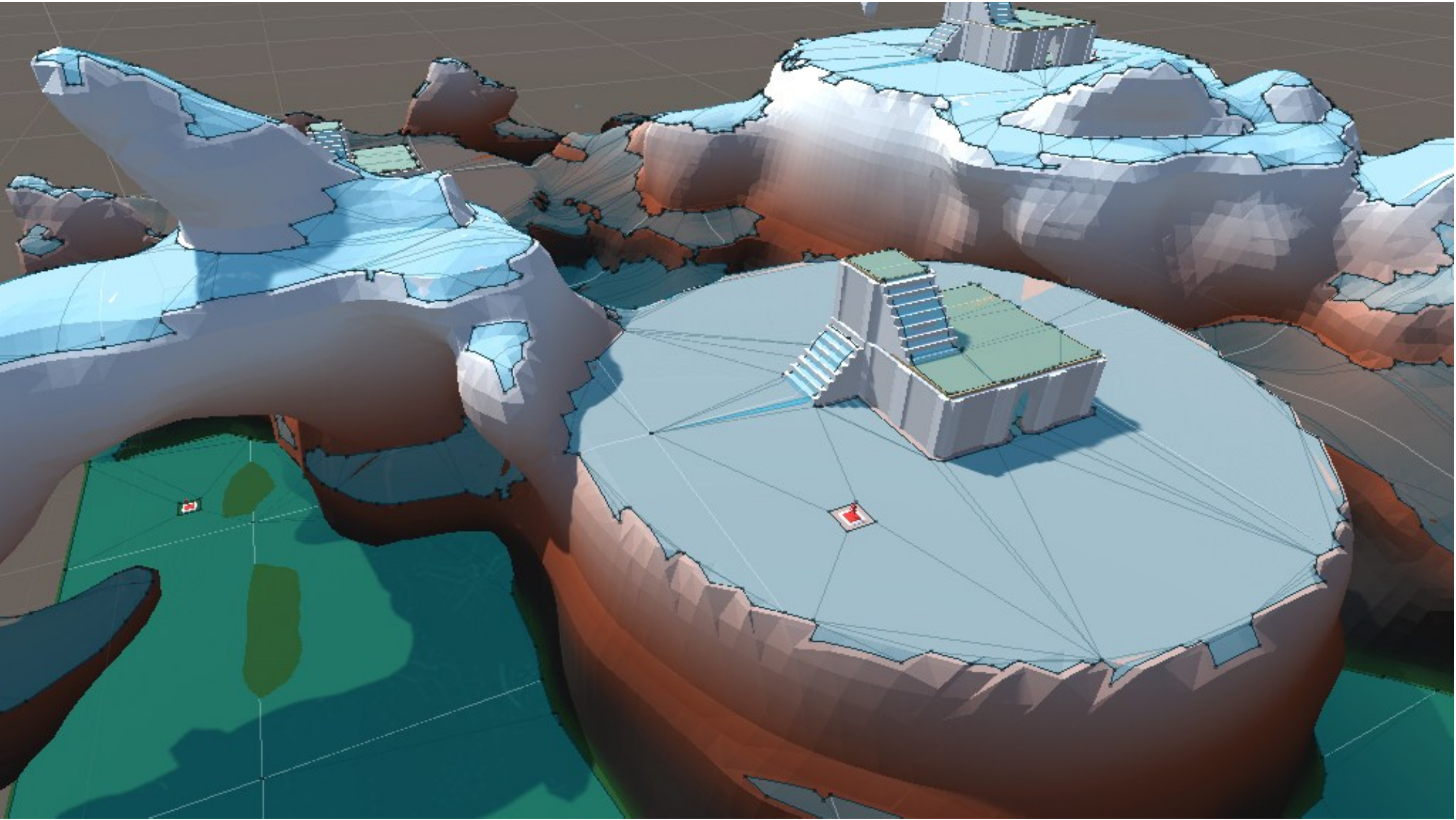}
    \caption{The proposed hybrid planner is capable of performing real-time navigation in complex 3D gaming environments featuring geometrically disconnected regions only reachable by agents through special actions. These environments are difficult for approaches based on the NavMesh.}
    \label{fig:nav_mesh}
    % \vskip -0.1in
\end{figure}

Currently, with few exceptions, navigation in video games relies on the NavMesh, which is created and used as follows.
\begin{enumerate}
    \item A graph representation of the world is pre-generated from the game geometry.
    \item At runtime a pathfinding algorithm such as A* \cite{hart1968aFormalBasis} is run on this graph to find the shortest path between two locations in the game.
    \item A player controller, attached to the character, is used for point based navigation to follow the sequence of way-points returned by the planner.
\end{enumerate}
While the navigation mesh provides a compact representation of the world, it is somewhat limited and results in disconnected regions in complex environments, as demonstrated in Figure \ref{fig:nav_mesh}. The disconnected regions are due to some assumptions of the underlying NavMesh generation algorithm. The user can provide the maximum inclination and height the agent can walk and jump, but more complex actions such as double jumping or using a jump pad are not taken into account. Connections due to special character abilities such as jump pads, grappling hooks and teleporters often involve time consuming manual editing of the graph in order to add additional links. While human intervention is possible, it is not feasible when the environments are procedurally generated as there can be a limitless number of new environment instances. Some automatic approaches do exist but are currently limited to simple actions \cite{axelrod_graphGenerationInDynamicWorlds,Roumimper2017MeshNavThroughJumping,Budde2013AutoGenJumpLinks}.

Recent works have looked to address this problem with an end-to-end Deep Reinforcement Learning (RL) approach \cite{ijcai2021-294} and have shown that when trained and evaluated on small (50m$\times$50m) to medium (100m$\times$100m) sized maps, end-to-end approaches achieve good performance. However, when evaluated in larger (300m$\times$300m) environments, policies learned using RL perform well on near and medium distance goals but poorly on farther challenging goals. Euclidean distance is not the best measure of complexity in this case, but as the odds of encountering a hard obstacle increase with the size of the map, it is a decent proxy for the complexity of navigation. 

%as it does not take into account the obstacles the agent must avoid in order to reach its goal. Distance is correlated with complexity, as it increases as the agent is likely to encounter more obstacles, needing to explore and backtrack. Thus, the sequence of actions the agent must output in order to reach the goal becomes larger.

In this paper we explore a hybrid approach that aims to combine the strengths of a classical graph-based NavMesh, with the capacity of end-to-end RL to incorporate complex abilities to improve navigation performance. Designed around the stringent constraints of online gaming environments, our method is able to generate new navigable graphs with low computational complexity.
The proposed graph augmented RL agent outperforms the generalization performance of a learning-based agent on a set of maps of 250m$\times$250m. In addition, the augmented approach also greatly improves the performance of a map specific agent that is trained and evaluated on a fixed game world of 1km$\times$1km, 10 times larger than those typically used in the research community \cite{ijcai2021-294}.

We also introduce the ``\emph{GameRLand3D}'' environment, a new benchmark, which pushes the size and complexity of gaming environments to new limits. It features realistic production-size maps as well as well as complex navigation tasks suited for modern 3D video games featuring navigation actions like jumps, pads, zip lines etc. We think this benchmark is well suited to quantify the performance of end-to-end approaches and to explore potential novel solutions. 

%The paper is structured as follows, we begin with a study of the related works in section \ref{sec:related_work}, we then describe the GameRLand3D environment and task definition in \ref{sec:environment}. In section \ref{sec:methods}, we define several graph methods that we have implemented, followed in section \ref{sec:experiments} by a detailed empirical evaluation of generalization performance, run-time performance (crucial in real time video games) and how well our approach scales to environments of 1 million $m^2$.

\vspace{1mm}
\noindent Our contributions are as follows:
\begin{itemize}
\item We quantify the limitations of the performance of end-to-end Deep RL approaches in complex and large-scale environments.
\item We propose a novel hybrid graph and RL approach that decouples long distances planning and low level navigation and outperforms end-to-end Deep RL methods in both a generalization and map specific setup. Notably, this approach provides interpretable plans which can be tweaked by a game designer in order to modify the agent behavior without the need to fully retrain it.
\item We release of an open-source 3D procedural environment which can generate complex 3D worlds of more than one million square meters. 
\item We introduce a detailed empirical evaluation and ablation of two hybrid RL and graph based approaches, with a special focus on their run time performance, a key requirement of video games operating in real time at 60 frames per second.
\end{itemize}

\section{Related work}\label{sec:related_work}
\subsection{Deep RL environments}
\begin{figure}
    \centering
    \includegraphics[width=0.5\textwidth]{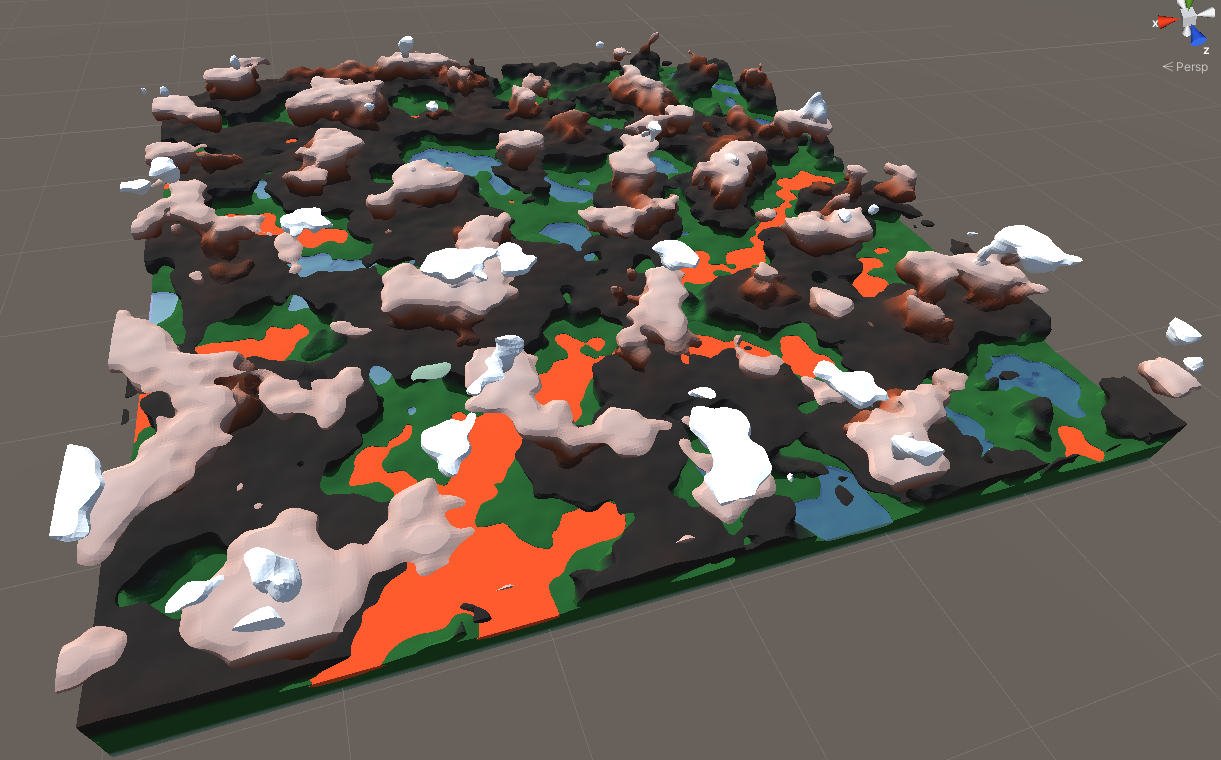}
    \caption{A vast, 1km by 1km procedurally generated map generated with in the GameRLand3D environment.}
    \label{fig:environment}
    % \vskip -0.1in
\end{figure}

There has been a proliferation of new Deep RL environments in the last five years, the most well known being the gym framework \cite{Brockman2016OpenAIGym} which provides an interface to fully observable games such as the Atari-57 benchmark and several continuous control tasks. There are a variety of partially observable 3D simulators available, starting with more game-like environments such as the ViZDoom simulator \cite{wydmuch2018vizdoom}, DeepMind-Lab \cite{Beattie2016DeepMindLab} and Malmo \cite{Johnson}. A number of photo-realistic simulators have also been released such as Gibson \cite{xiazamirhe2018gibsonenv} and AI2Thor \cite{KolveAI2-THOR:AI}. The highly efficient Habitat-Lab simulator \cite{habitat19iccv} builds on the datasets of Gibson and AI2Thor and implements a number of benchmark 3D navigation tasks. The aforementioned environments were designed to train and evaluate RL agents in relatively small environments of the order of hundreds of square metres, with relatively small action spaces. Moreover, they do not consider the complex actions that are in modern video games, such as double jumping, teleportation, jump pads, grappling hooks, and wall runs. The objective of the GameRLand3D environment, shown in Figure \ref{fig:environment},  is to provide a realistic benchmark for real world video game environments, to identify the limitations of current learning-based techniques and to allow reasearcher to explore novel alternatives.
\subsection{Classical planning and navigation in video games}
Existing methods for path planning in the video game industry mostly rely on building a Navigation Mesh (NavMesh)\cite{snook2000navmesh}. A NavMesh is a 2D/3D graph whose nodes represent convex polygonal regions of the walkable space, typically triangles, and whose edges indicate possibilities of navigation between regions. To find a path between two locations in the world, one can then apply path finding algorithms like A* \cite{Hart1968} or one of its variants on this graph. This classical approach is robust for many applications but presents some limitations. In particular, when it is possible for the agent to use mobility actions (like dashing, jumping, double jumping, using a jetpack), the high number of places it can reach from each location dramatically increases the connectivity of the NavMesh \cite{ijcai2021-294, gordillo2021improving}. Therefore, it increases its memory cost as well as the runtime cost of the pathfinding algorithms, to the point where using a NavMesh ends up being too costly, constraining game design. Another important aspect that a NavMesh does not account for is the kind of constraint that a character can have, such as the turn radius of a car or a plane in 3D. This kind of constraint can be impossible to encode in a NavMesh in an efficient manner. 

Contrary to NavMesh-based approaches, RL for point-to-point navigation merges the path finding and the character controller (code the agent actually uses to move and follow the path) components by directly outputting actions the agent needs to take at every step to go from point A to point B. This makes using RL especially attractive since it could compress the semantics/topology of the map in a much more efficient way while having a fixed low cost to run. % NavMesh based approaches have also been extended to dynamic environments \cite{https://doi.org/10.1002/cav.1468}.

\subsection{Neural approaches to navigation}
The field of Deep Reinforcement Learning (RL) has gained attention with successes on board games \cite{silver2016mastering} and Atari games \cite{mnih2015human}. Recent works have applied Deep RL for the control of an agent in 3D environments \cite{mirowski2016learning,jaderberg2017unreal}, exploring the use of auxiliary tasks such as depth prediction, loop detection and reward prediction to accelerate learning. Other recent work uses street-view scenes to train an agent to navigate in city environments \cite{mirowski2018learning}. 

To infer long term dependencies and store pertinent information about the partially observable environment, network architectures typically incorporate recurrent memory such as Gated Recurrent Units \cite{chung2015gated} or Long Short-Term Memory \cite{hochreiter1997long}. Extensions to memory based neural approaches began with Neural Turing Machines \cite{graves2014neural} and Differentiable Neural Computers \cite{graves2016hybrid} and have been adapted to Deep RL agents \cite{wayne2018unsupervised}. 

Spatially structured memory architectures have been shown to augment an agent’s performance in 3D environments and are broadly split into two categories: metric maps which discretize the environment into a grid based structure and topological maps which produce node embeddings at key points in the environment. Research in learning to use a metric map is extensive and includes spatially structured memory \cite{parisotto2017neural}, Neural SLAM based approaches \cite{zhang2017neural} and approaches incorporating projective geometry and neural memory \cite{gupta2017cognitive,bhatti2016playing}. These techniques are combined, extended and evaluated in \cite{beeching2020egomap, NEURIPS2020_6e01383f}. Research combining learning, navigation in 3D environments and topological representations has been limited in recent years with notable works being \cite{savinov2018semi} who create a graph through random exploration in the ViZDoom RL environment \cite{wydmuch2018vizdoom}. \cite{eysenbach2019search} also performs planning in 3D environments on a graph-based structure created from randomly sampled observations, with node distances estimated with value estimates. \cite{beeching2020learning} perform topological planning with a neural approximation of a classical planning algorithm that can be applied in uncertain environments. Other recent approaches such as \cite{Chaplot_2020_CVPR} build a graph structure, but rely on 360 degree camera measurements. Practically all of the aforementioned approaches are applied in small planar environments of hundreds of square metres, and scaling these methods to the environment where we apply our hybrid approach poses several challenges. Availability of ground truth: most graph based methods require ground truth actions and graph connectivity, as the calculation of optimal actions is often intractable in 3D environments of hundreds of thousands of square metres this precludes this possibility. We are also constrained due to inference time: many of these methods do not operate at real time speed, particularly when controlling hundreds of agents in parallel. 
\begin{figure}
    \centering
    \includegraphics[width=0.5\textwidth]{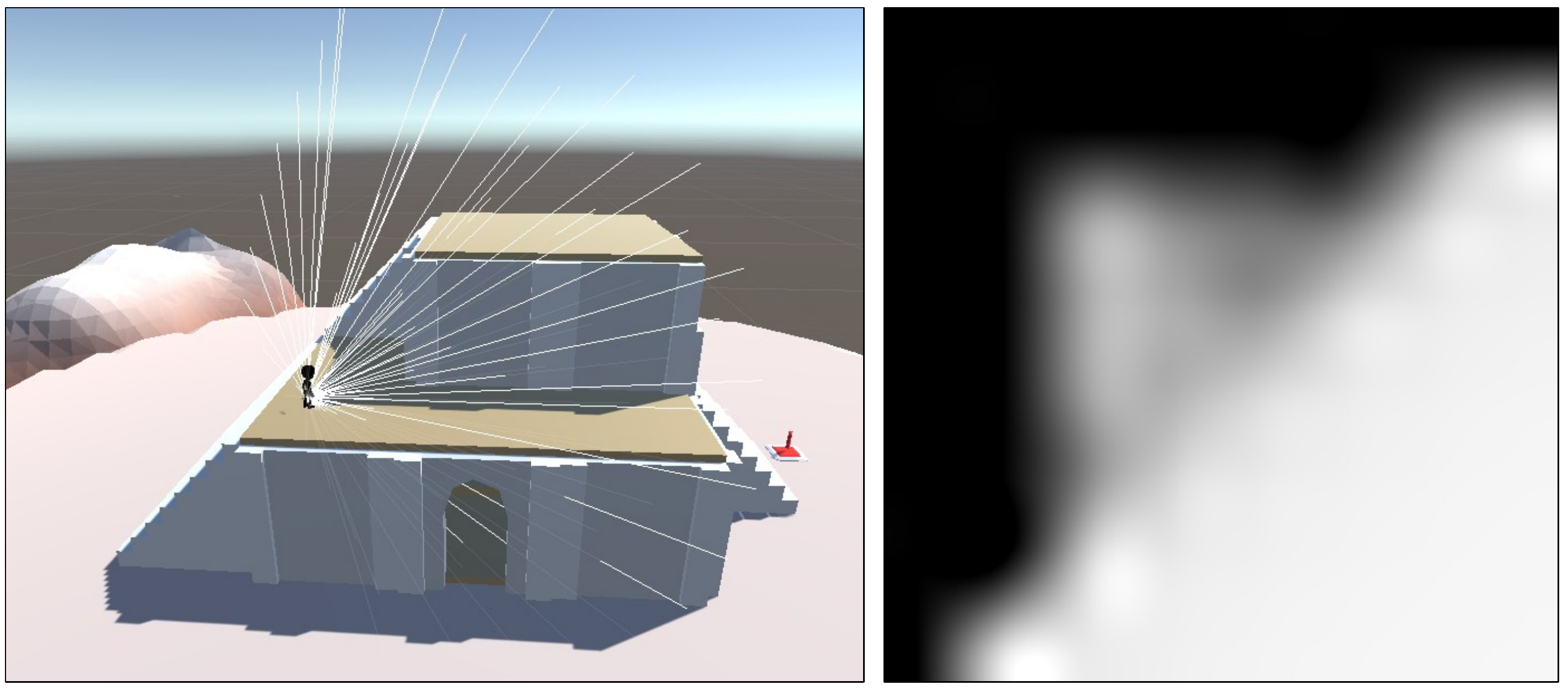}
    \caption{Left: An agent's depth observation in the GameRLand3D environment. Right: the resulting 2D observation in tensor form, that will be processed by the agent's CNN.}
    \label{fig:agent_observation}
\end{figure}
\section{Environment and task definition}\label{sec:environment}
In order to address the challenging problem of navigation in vast complex 3D environments we have developed the GameRLand3D environment. GameRLand3D allows the creation of enormous worlds and enables complex interactions such as double jumping and jump-pads, includes buildings, water and lava hazards. The environment is built using the Unity-ML agents \cite{juliani2018unity} framework, which provides a versatile tool to build RL environments. We interact with the environment by controlling an agent, represented as a 1.8 meter humanoid character, shown in Figure \ref{fig:agent_observation}. 

Our environment implementation uses 3D Perlin noise \cite{perlin1985image} for the procedural generation of the game world, followed by random placements of buildings, plateaus and jump pads to create a challenging state space for agents to interact with, explore and navigate.

\subsubsection{Task}
The task we consider is point-to-point navigation where the agent is spawned at a location  $p_{spawn}$ and must navigate to a goal location $p_{goal}$. Start and goal locations are sample from navigable terrain, using a NavMesh for simplicity, through a function $f_{navmesh}$ from which we can sample valid navigation locations. The NavMesh contains many disconnected regions and cannot be directly used for long distances navigation, but can be used to identify valid spawn locations. To reduce overall inference time and to allow for planning over longer time horizons, we request a decision from the agent's policy network every 10 time-steps, repeating the actions for the intermediary 10 steps. % move to agent section, this is an RL choice not a env choice
\subsubsection{Constraints}
Video game environments provide certain advantages for training RL agents when compared to learning a model that may be transferred to the real world. The pose of the agent is known and we are not expected to rely on noisy pose estimates and fulfill safety guarantees, which add an additional challenge to robotic navigation. Video games do have other requirements and constraints, model implementations must be highly optimized, parallelized and run at real time speeds, making decisions at a minimum of 60 times per second. If an agent is player facing, its behavior must be credibly "human-like", a subjective metric that is discussed in \cite{devlin2021navigation}. Ideally, methods should provide interpretable information, so that game designers can be confident about adopting it in production.
\subsubsection{Observations}
Rendering agent observations from the environment with a virtual monocular camera is prohibitively expensive for NPC control in video games, so the agent's observations are a depth measurement of a $8{\times}8$ cone of raycasts in front of the agent (see Figure \ref{fig:agent_observation}). We also provide useful information such as the location of the goal in the agent's frame of reference, the agent's velocity and acceleration. 

\subsubsection{Actions}
The environment actions are continuous and correspond to actions that are typically available to a human player: forward, backward, strafe, turn, jump and double-jump (where a player can execute a second jump while in mid-air). The jump action is treated like a continuous action for the RL policy and is discretized in the environment.
\subsubsection{Rewards}
Building on the work of \cite{ijcai2021-294}, we combine three rewards: A \textbf{sparse reward} of +1 when the agent succeeds in the task and -1 when the agent fails. A \textbf{dense reward} every step in which the agent reduces its best euclidean distance to the goal location, divided by a normalization factor $\lambda$. A \textbf{time-step penalty} of -0.0005 per step, to encourage the agent to perform the task quickly.
%\subsubsection{Episode termination condition}
An episode is terminated when the agent is within 1m of its goal, the agent falls off the map or the agent has not decreased its best euclidean distance to the goal for 300 time-steps.
\begin{figure}
    \centering
    \includegraphics[width=0.50\textwidth]{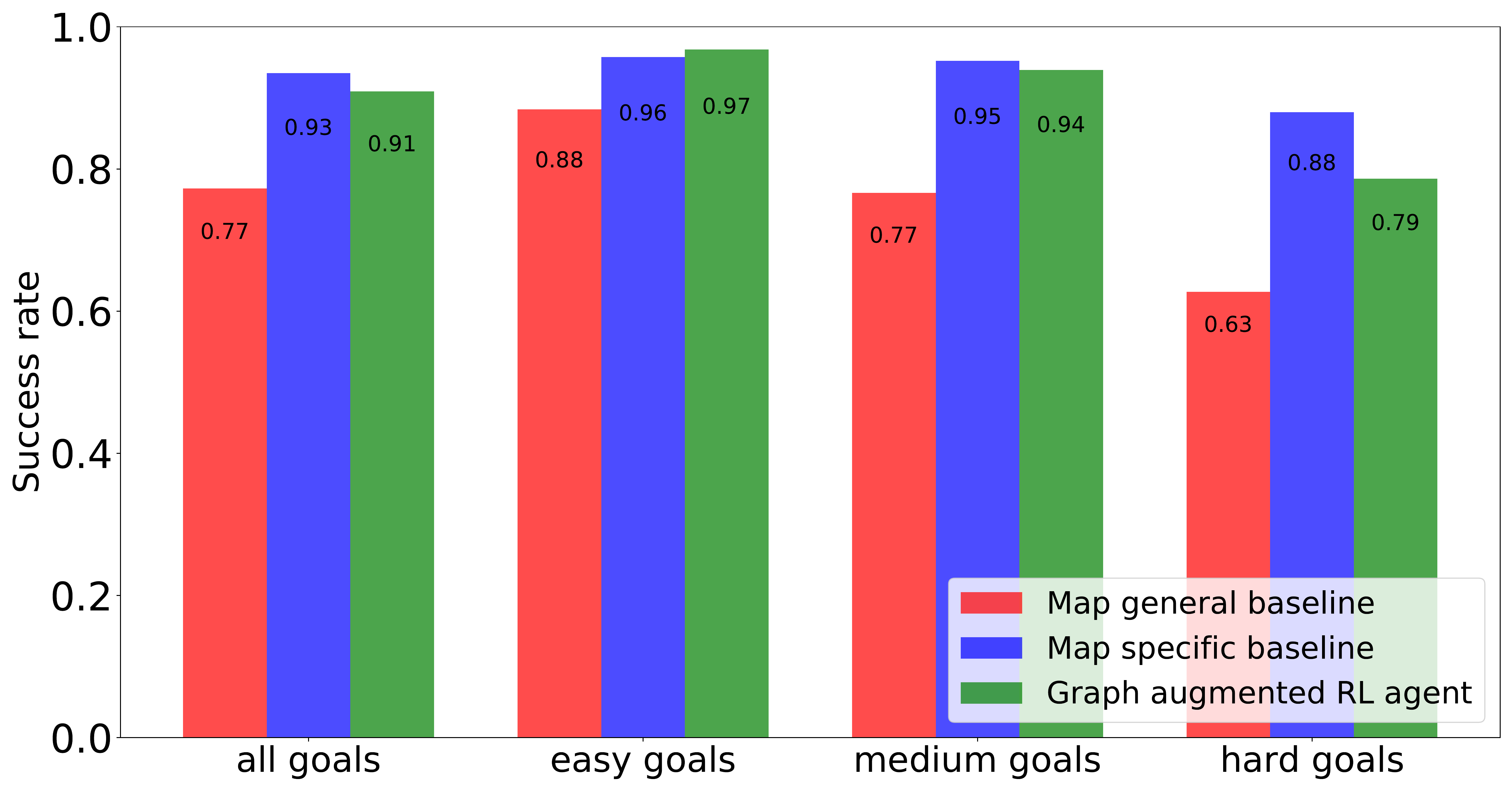}
    \caption{The generalization gap in performance between a map specific recurrent agent trained on a single map and a general recurrent agent trained on 128 maps and then evaluated on the same unseen map. Compared against a Graph augmented RL agents, introduced in this paper.}
    \label{fig:generalization_gap}
    % \vskip -0.1in
\end{figure}
\begin{figure*}
    \centering
    \includegraphics[width=1.0\textwidth]{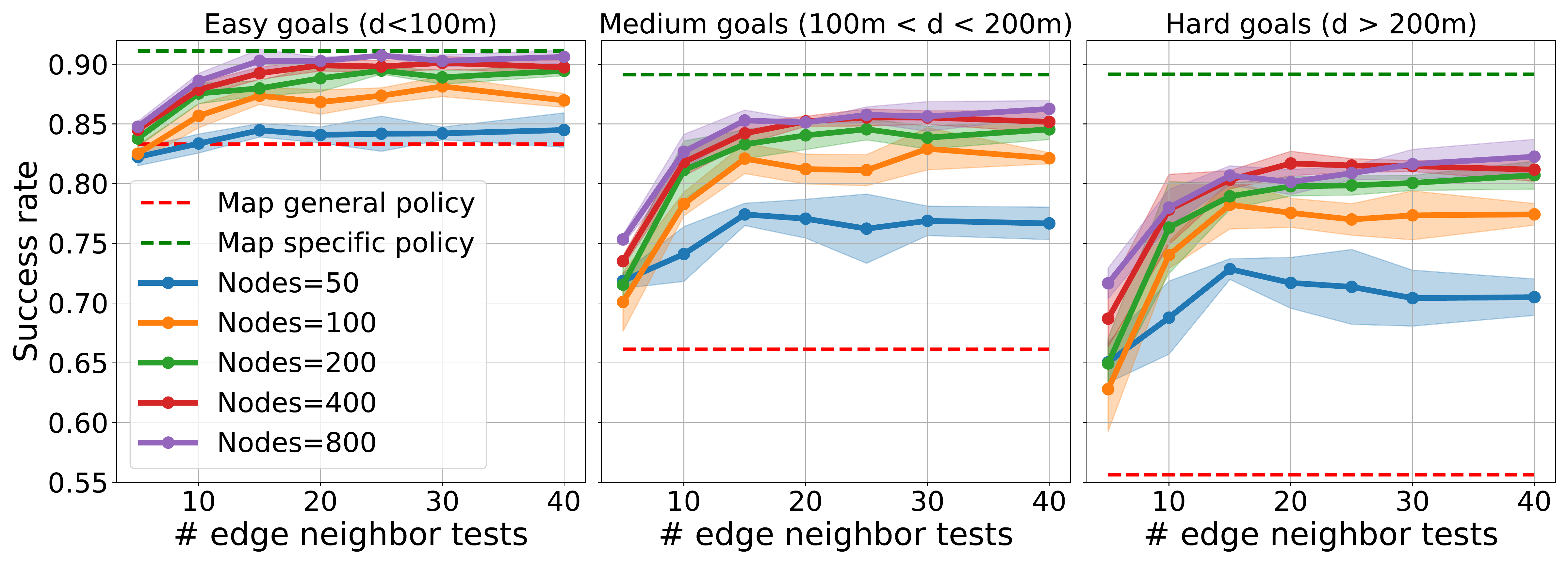}
    \caption{Results for different hyper-parameters for the method \emph{Distance constrained random vertex placement}: graph sizes of 50 to 800 nodes; edge candidates of 5 to 40 neighbors}
    \label{fig:random_method_coverage}
    % \vskip -0.1in
\end{figure*}
\section{Empirical evaluation of end-to-end methods}
We initially trained a baseline policy, in order to quantify the empirical performance of end-to-end Deep RL approaches in such large-scale worlds. In particular, we were interested in generalization, how well a policy trained in one set of environments transfers to a new environment. The baseline policy was trained using a recurrent version of a sample efficient off-policy actor critic RL algorithm, Soft Actor Critic (SAC) \cite{haarnoja2018soft}, with a learned entropy coefficient and no state value network \cite{haarnoja2018SACApplications}. The SAC algorithm aims to finds a policy that maximizes cumulative reward and the entropy of each visited state.

\begin{equation}
\pi^{*}=\arg \max_{\pi} \sum_{t}\mathbb{E}_{(\mathbf{s}_{t}, \mathbf{a}_{t}) \sim \rho_{\pi}}  [r(\mathbf{s}_{t}, \mathbf{a}_{t})+\alpha \mathcal{H}(\pi(\cdot \mid \mathbf{s}_{t}))]
\label{eq:sac_entropy}
\end{equation}
%
% \begin{equation}
% \pi^{*}=\arg \max_{\pi} \sum_{t} \mathbb{E}_{\left(\mathbf{s}_{t}, \mathbf{a}_{t}\right) \sim \rho_{\pi}}\left[r\left(\mathbf{s}_{t}, \mathbf{a}_{t}\right)+\alpha \mathcal{H}\left(\pi\left(\cdot \mid \mathbf{s}_{t}\right)\right)\right]
% \label{eq:sac_entropy}
% \end{equation}

We include a learnable entropy temperature parameter $\tau$, to enable automatic optimization of the entropy of the objective function. We found that having independent policy and Q-network embeddings improves the reliability of the Q-value estimates and led to more robust, stable training. We trained several versions of the baseline agent architecture. A \textbf{reactive} agent that receives the 8x8 raycast observation and a \textbf{memory-based} recurrent agent that extends the reactive agent with an LSTM in order to be able to learn longer term dependencies. We use a burn-in of 20 time-steps to initialize the hidden state as recommended in \cite{kapturowski2018recurrent, paine2019making}.

In order to quantify the reduction in performance when a policy is learned on one specific map or when a policy is learned on a set of maps and evaluated on another, we trained two versions of the memory based agent: a general agent and a map specific agent. The general agent is trained on a large dataset of 128 procedurally generated map instances. During training we evaluate the performance of the policy on a set of 4 validation maps and keep the model with the highest success rate. In addition, independently we train 4 map specific agents, one per test map, the objective being to quantify the decrease in performance, or \textit{generalization gap}, between the map specific and general agent, see Figure \ref{fig:generalization_gap}. The graph augmented methods that we discuss in the following section use the general baseline for point-to-point navigation and achieve comparable performance to the map specific agent. We note that a general agent is a crucial requirement in the video game production setting, as it removes the need of retraining on every new map instance.

\section{Graph augmented RL agents}\label{sec:methods}
Pure end-to-end training of vanilla agents with recurrent memory leads to sub-optimal solutions for large and complex environments, as planning is performed on memory without spatial structure. While spatial organization of neural memory can be learned implicitly in principle, as shown by the emergence of grid-cells in unstructured agents trained for localization \cite{GridCellsICLR2018}, in practice this remains a hard task, in particular from reward alone. On the other hand, and as mentioned, the NavMesh alone is hardly applicable in the presence of special agent actions leading to disconnected regions.

We address this problem and propose a hybrid method which combines a directed graph (digraph) for planning and way-point to way-point navigation with a low-level policy trained with RL. This approach decomposes the problem of navigation over long distances into sequences of smaller sub-problems. The motivation is that (i) small-scale way-point navigation can be learned easily through reward, even in the presence of disconnected regions accessible with jump actions; (ii) large-scale navigation in a connected graph can be solved efficiently with optimal symbolic algorithms, like Dijkstra's algorithm or A*. 
%By breaking down a long distance goal into a sequence of easy, short-term goals, 
We will show that the proposed method features a high success rate, provides interpretable paths and allows for identification of disconnected regions in the environment topology.

We explored several variants of the proposed method, which differ in the way how they dynamically create graphs when a new environment is spawned. For all variants, we build a graph $G$ comprised of a set of vertices $V$ and edges $E$, where edge $e(i,j)$ connects vertex $v_i$ to vertex $v_j$. Each edge has a cost $c(i,j)$. Assigned to each vertex is a 3D position in the game. Planning from a position $p_{start}$ to a goal position $p_{goal}$ is performed by identifying the nearest vertex in the graph to the starting position $v_{start}$ and the nearest vertex to the goal position $v_{goal}$. The optimal sequence of vertex locations to traverse in order to reach $v_{goal}$ is found with the classical path planning algorithm Dijkstra \cite{dijkstra1959note}.

The process of graph creation follows three key steps: identify vertex locations, testing edge connectivity and graph building. The following sections describe the different methods we explored for identifying the vertex locations, edges, edge costs and building the graph. %Each method provides a compromise between run-time performance, graph construction time and overall success rate. As user requirements may differ from one game to the next, we provide analysis of each approach in the experiments section \ref{sec:experiments}.

\subsection{Unconstrained random vertex placement}
In the unconstrained random approach we sample $n$ vertices from the function $f_{navmesh}$. This provides vertices that are located on navigable terrain, but the edge connectivity is unknown. Edge connectivity is computed by first identifying the $k$ nearest neighbors of each vertex and then evaluating a short point-goal episode between each vertex and its k-nearest neighbors. If the episode is successful then the directed edge is considered to be connected, with its cost equal to the number of steps in the episode. Using the agent steps as a cost over the euclidean distance leads to graph planning that takes into account local obstacles in the planning process.

\subsection{Distance constrained random vertex placement}
We augment the unconstrained random approach with coverage criteria, in order to spread nodes evenly across the environment and ensure that we can estimate connectivity all over the map. We aim to identify the set of vertex positions $V$ that satisfy the criteria $d_{V}:=\max _{x, y \in V, x \neq y} d(x, y)$, where $d(x,y)$ is the euclidean distance between two vertex positions. We find the set of positions $V$ using a greedy strategy. We first calculating the optimal average distance between vertices, assuming the environment was a planar surface, and then sampling vertex locations from $f_{navmesh}$ until no nodes are present in the proximity of the sampled vertex location. In order to avoid infinite loops this test is repeated 100 times and after each test the distance criterion is decayed by a factor of 0.98.

\begin{table*}[ht]
\centering
\label{tab:all_results}
%\resizebox{\textwidth}{!}{%
{ \small 
\begin{tabular}{l|llll}
\hline \hline
                                       & \multicolumn{4}{c}{\textbf{Success rate}}                                     \\ 
\multicolumn{1}{c|}{\textbf{Method}} &
  \multicolumn{1}{c}{\textbf{All goals}} &
  \multicolumn{1}{c}{\textbf{Easy goals}} &
  \multicolumn{1}{c}{\textbf{Medium goals}} &
  \multicolumn{1}{c}{\textbf{Hard goals}} \\ \hline
Map specific LSTM baseline (not comparable)             & 0.899 $\pm$ 0.000 & 0.911 $\pm$ 0.000 & 0.891 $\pm$ 0.001 & 0.892 $\pm$ 0.001 \\ \hline
General reactive baseline              & 0.680 $\pm$ 0.000 & 0.821 $\pm$ 0.000 & 0.619 $\pm$ 0.000 & 0.574 $\pm$ 0.000 \\
General LSTM baseline                  & 0.696 $\pm$ 0.000 & 0.833 $\pm$ 0.000 & 0.661 $\pm$ 0.001 & 0.556 $\pm$ 0.002 \\
Unconstrained random vertex placement                  & 0.824 $\pm$ 0.023 & 0.878 $\pm$ 0.013 & 0.812 $\pm$ 0.029 & 0.767 $\pm$ 0.029 \\
Distance constrained random vertex placement    & 0.859 $\pm$ 0.002 & 0.904 $\pm$ 0.002 & 0.850 $\pm$ 0.004 & 0.809 $\pm$ 0.004 \\
% Density based vertex placement                    & 0.833 $\pm$ 0.004 & 0.888 $\pm$ 0.007 & 0.820 $\pm$ 0.004 & 0.776 $\pm$ 0.006 \\
% Trajectory merging                          & 0.812 $\pm$ 0.010 & 0.880 $\pm$ 0.010 & 0.808 $\pm$ 0.010 & 0.724 $\pm$ 0.009 \\
% Trajectory merging vertices with edge tests & 0.862 $\pm$ 0.002 & 0.903 $\pm$ 0.004 & 0.855 $\pm$ 0.001 & 0.816 $\pm$ 0.002 \\
% Flood                                  & 0.800 $\pm$ 0.027 & 0.822 $\pm$ 0.040 & 0.804 $\pm$ 0.023 & 0.765 $\pm$ 0.020 \\
% Flood with grandchild edges            & 0.856 $\pm$ 0.008 & 0.898 $\pm$ 0.004 & 0.849 $\pm$ 0.012 & 0.808 $\pm$ 0.010 \\ 
\hline \hline
\end{tabular}%
}
\caption{Evaluation on a test set of 4 maps of size 250m${\times}$250m. In addition to general reach and memory-based policies, we include the performance of 4 map specific policies, learned independently on each test map. We compare against the best performance of the unconstrained and constrained coverage-based hybrid methods.}
\label{tab:results}
\end{table*}

\subsection{Optimizations}
\subsubsection{Robust estimation of edge connectivity}\label{sec:edge_repeats}
One weakness of performing the edge connectivity test once is that there is a small probability that agents can introduce an edge in the graph that often fails, leading to  overconfident planning. This problem is compounded when there are several edges in the graph that are difficult to navigate. In order to identify and prune these edges, we evaluate an option to repeat the edge connectivity tests several times and keep the edges where the percentage of the tests exceed a given threshold. This approach leads to robust plans and higher performance, with the downside of a longer graph building step.

\subsubsection{Re-planning}\label{sec:edge_repeats}
Qualitative analysis demonstrated that occasionally a graph augmented agent would fail to reach its next way-point, resulting in an agent that was often stuck as it had fallen off the side of an obstacle. This was resolved with the implementation of a re-planning step, where if the agent does not decrease its distance to the next way-point for 50 time-steps we automatically re-plan from its current location to the goal location. We observed that re-planning increased the success rate of all the examined methods.

\section{Experiments}\label{sec:experiments}
\subsubsection{Setup}
We generate maps of size 250m$\times$250m$\times$80m, 128 training maps,  4 validation maps and 4 tests maps to test the baseline and graph augmented methods. We then scale this approach to maps of 1km$\times$1km $\times$80m to evaluate the performance in vast game worlds. Such a map is notably larger than the largest maps tested in the literature \cite{ijcai2021-294}.

 While other works also include the Success weighted by Path Length (SPL) metric \cite{anderson2018evaluation}, we are unable to compare against this metric as the SPL metric. As SPL requires ground truth shortest paths, which cannot be computed in such large environments with complex action spaces. We separate the goals into three groups, easy, medium and hard, based on distance between the start and end points to break down the performance of each method. The results on hard, long distance goals, are of particular interest as this is where the baseline policy had the worst performance. 

\subsubsection{Hyper-parameter optimizations}
We extensively optimized the hyper-parameters of each approach and report the results of the best performing configuration for each technique. Each test was evaluated with three seeds in order to quantify the average and variation of performance that can be achieved with these methods. We evaluate the performance of each method using the success rate metric on 4 test maps, with 2,500 goals per map.
As an example, we report in Figure \ref{fig:random_method_coverage} the evaluation of a single method, \emph{Random vertex with coverage constraint}, in numerous hyper-parameter configurations. In this case we have varied the number of nodes and the number of nearest neighbors with which to perform edge connectivity tests.

\subsubsection{Comparisons}
We evaluate the performance of the proposed graph augmented methods and compare them to the performance of the recurrent baselines in table \ref{tab:results}. We observe that the \emph{Distance Constrained Random Vertex Placement} outperforms the baseline method with a success rate of 85.9\%, an improvement of 16.3\%. We also implemented flooding and trajectory based methods which are excluded for brevity.%, but are discussed in the overview video \href{https://youtu.be/H0WAHvEeVyc}.

\subsubsection{Run-time analysis}
\begin{figure}[t]
    \centering
    \includegraphics[width=0.50\textwidth]{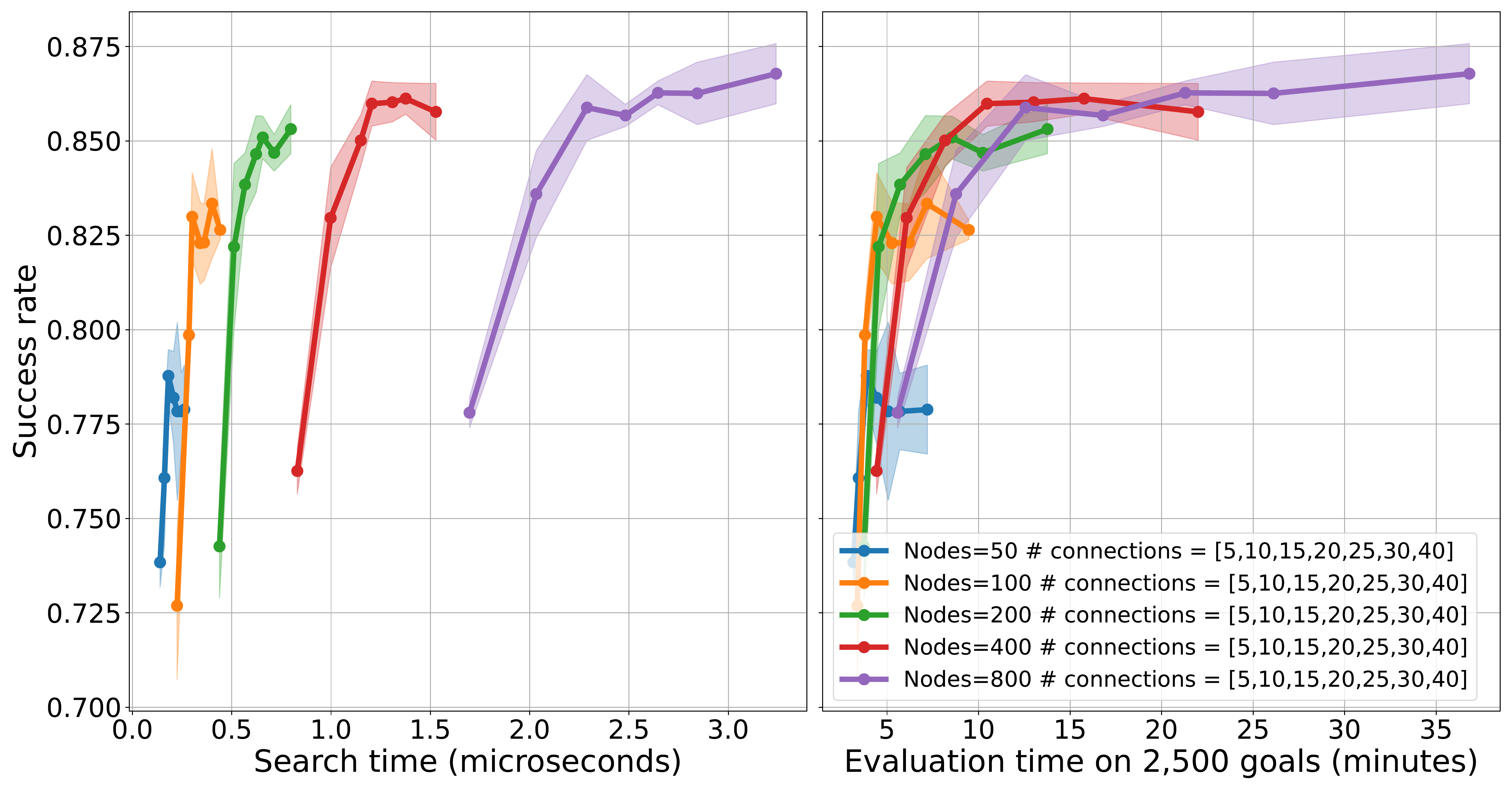}
    \caption{Run-time trade-offs of the coverage constraint approach. Left: Average graph search time in ms in several configurations. Right: Average graph construction and evaluation time in minutes on 2,500 goals.}
    \label{fig:performance_analysis}
    % \vskip -0.1in
\end{figure}

Run-time performance is critical in real time video games, where the game logic and rendering must be executed 60 times per second. We perform an analysis of the run-time performance of the various approaches. In particular, we are interested in the time it takes to build the graph, to evaluate 2,500 goals on a test map and average duration of a graph search when exploiting the hybrid approach. Figure \ref{fig:performance_analysis} summarizes the results for key sets of hyper-parameters for the \emph{Distance Constrained Random Vertex Placement} approach. While larger graphs improve the performance of the augmented agent, they also increase the evaluation time and the time taken to perform a graph search. There is clearly a trade-off to be made, with the optimal choice depending on the exact use case, for an offline exhaustive test of a map instance a higher node count is suitable. Whereas if the graph must be created in near real-time a smaller node count would be required.

%We observe that there is a trade-off, with graph trajectory based methods being faster to build, but achieving poorer performance. The optimal choice therefore depends on the exact use case --- for an offline exhaustive test of a map, the longer running random or flood methods would be more suitable, whereas if the use-case requires a graph to be built in near real time, the trajectory based methods may be more suitable. 

\subsubsection{Scaling to production size maps}
\begin{figure}
\centering
    \includegraphics[width=0.45\textwidth]{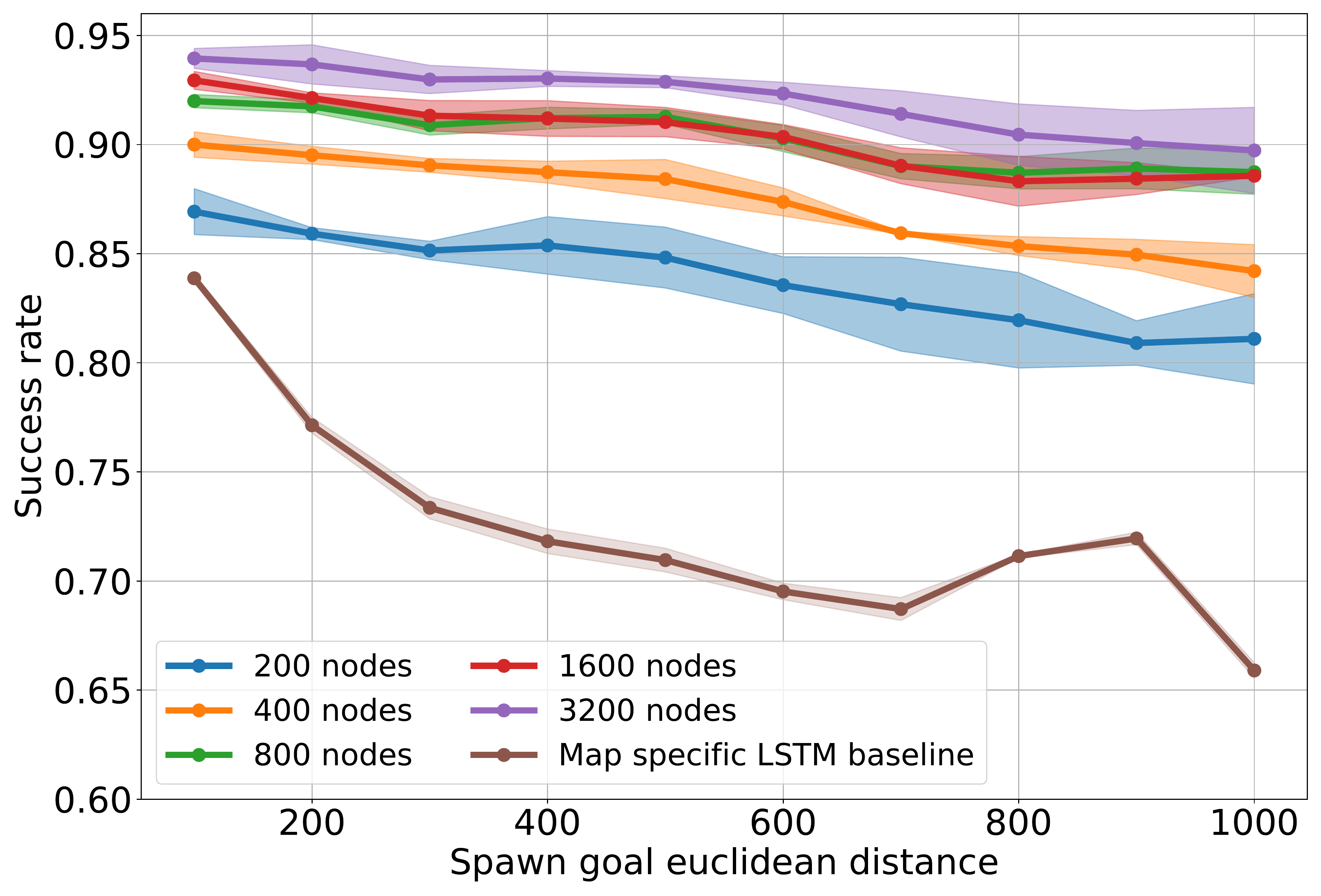}
    \caption{Evaluation of the Random Graph with coverage approach on a single $1km^2$ map with 10,000 goals. Compared against a map specific LSTM baseline policy. Shown is goal distance, in 100m bins, compared with the success rate of each approach. We observe that the method begins to reach diminishing returns at 1,600 vertex locations. Overall average success rate is increased from 72\% to 92\%.}
\label{fig:results_big_map}
\end{figure}

Previous experiments have focused on generalization of a learned policy to unseen maps. We have extended our method to a vast 1km$\times$1km$\times$200m map. We find that even a map specific baseline RL agent, whose policy is learned directly on the map it is evaluated on (environment overfit), performs poorly compared to our hybrid agent, with an overall success rate of 72\%. We augment the map specific agent with the \emph{Distance Constrained Random Vertex Placement approach} and evaluated the performance on 10,000 goals with graphs of size 200 to 1,600 nodes. We achieve an increase in success rate of 20\% on all goals, shown in Figure \ref{fig:results_big_map}.

% \begin{figure}
% \begin{subfigure}{.18\textwidth}
%   \centering
%   \includegraphics[width=1.0\linewidth]{figures/big_map_results_bar_plot.pdf}
%   %\caption{Success rate on all goals for baseline and ra, the graph based method achieves up to 20\% increased performance over and end-to-end approach.}
%   %\label{fig:sfig1}
% \end{subfigure}%
% \begin{subfigure}{.32\textwidth}
%   \centering
%   \includegraphics[width=1.0\linewidth]{figures/big_map_results_distance.pdf}
%   %\caption{Success rate compared against goal distances, goals binned into groups every 100m.}
%  % \label{fig:results_big_map}
% \end{subfigure}
% \caption{Evaluation of the Random Graph with coverage approach on a single $1km^2$ map with 10,000 goals. Compared against a map specific LSTM baseline policy. Left: performance on all goals of the LSTM baseline and graph augmented agents with a range of node densities, performance plateaus at 1600 nodes. Right: Goal distance, in 100m bins, compares with the success rate of each approach.}
% \label{fig:results_big_map}
% \end{figure}

\section{Conclusions}
We have introduced a set of graph augmented RL approaches for planning and navigation in vast 3D video game worlds. To demonstrate the capacity of these approaches we have developed the GameRLand3D environment, an open-source reinforcement learning environment. GameRLand3D enables procedural generation of vast open world environments with hazards such as water and lava, and features such as jump pads, buildings and overhangs. We have identified that end-to-end Deep RL approaches under-perform in large-scale environments. We evaluated two graph augmented RL techniques and performed extensive hyper-parameter tests to find a reasonable compromise between graph building time, run-time performance and success rate. Our hybrid approaches boost generalization performance from 69.6\% to 86.2\%. We also achieve a 20\% increase in success rate in enormous environments.

In addition to support and maintenance of the GameRLand3D environment, our roadmap includes the addition of more complex building types, player abilities such as teleportation and jetpacks, and other features such as ravines and caves. Future hybrid approaches in this environment will investigate the possibility of online graph building, using function approximation to predict edge connectivity in dynamic environments where the graph must be reconstructed when there is a change in map topology. 

\bibliography{aaai22}
\end{document}